\documentclass[titlepage]{article}

\usepackage{arxiv}
\usepackage[affil-it]{authblk}
\usepackage[utf8]{inputenc} 
\usepackage[T1]{fontenc}    
\usepackage[colorlinks=false, citecolor=black, hidelinks=true]{hyperref}       
\usepackage{url}            
\usepackage{booktabs}       
\usepackage{amsfonts}       
\usepackage{nicefrac}       
\usepackage{microtype}      
\usepackage{lipsum}
\usepackage{graphicx}
\graphicspath{ {./images/} }

\title{Visualizing the Finer Cluster Structure of Large-Scale and High-Dimensional Data}

\author[*] {Yu Liang}
\author[ ] {Arin Chaudhuri}
\author[ ] {Haoyu Wang}
\affil[ ] {IoT Analytics, SAS Inc.} 
\affil[ ] {701 SAS Campus Drive, Cary, NC 27519}
\affil[*]{Corresponding author: Yu Liang, yu.liang@sas.com}

\begin{document}
\maketitle

\begin{abstract}
Dimension reduction and visualization of high-dimensional data have
become very important research topics because of the rapid growth of large databases in data science. 
In this paper, we propose using a generalized sigmoid function to model
the distance similarity in both high- and low-dimensional spaces.
In particular, the parameter $b$ is introduced to the generalized
sigmoid function in low-dimensional space, so that we can adjust the 
heaviness of the function tail by changing the value of $b$.
Using both simulated and real-world data sets, we show
that our proposed method can generate visualization results
comparable to those of uniform manifold approximation and projection (UMAP), 
which is a newly developed manifold learning technique with fast running
speed, better global structure, and scalability to massive data sets. 
In addition, according to the purpose of the study and the data
structure, we can decrease or increase the value of $b$ to either reveal
the finer cluster structure of the data or maintain the neighborhood
continuity of the embedding for better visualization. Finally, we use domain knowledge to
demonstrate that the finer subclusters revealed with small values of $b$
are meaningful.
\end{abstract}

\keywords{Dimension Reduction \and Manifold Learning \and Data Visualization}


\section{Introduction}
Dimension reduction and visualization of high-dimensional data have
become very important research topics in many scientific fields because
of the rapid growth of data sets with large sample size and/or dimensions. 
 
In the literature of dimension reduction and information visualization, linear methods such as
principal component analysis (PCA) \cite{hotelling1933analysis} and
classical scaling \cite{torgerson1952first} mainly focus on preserving the most significant
structure or maximum variance in data; nonlinear methods
such as multidimensional scaling \cite{borg2005modern}, isomap
\cite{tenenbaum2000global}, and curvilinear component analysis (CCA) 
\cite{demartines1997curvilinear} mainly focus on preserving the 
long or short distances in the high-dimensional space. 
They generally perform well in preserving the global structure of
data but can fail to preserve the local structure. In 
recent years, the manifold learning methods, such as
SNE \cite{hinton2003stochastic}, Laplacian eigenmap
\cite{belkin2002laplacian}, LINE \cite{tang2015line}, LARGEVIS
\cite{tang2016visualizing}, t-SNE \cite{maaten2008visualizing}
\cite{van2014accelerating}, and UMAP \cite{mcinnes2018umap}, have gained
popularity because of their ability to preserve both the local 
and some aspects of the global structure of data. These methods generally assume that data 
lie on a low-dimensional manifold of the high-dimensional input space. 
They seek to find the manifold that preserves the 
intrinsic structure of the high-dimensional data. 

Many of the manifold learning methods suffer from something called the ``crowding
problem'' while preserving local distance of high-dimensional data in
low-dimensional 
space. This means that, if you want to describe small distances 
in high-dimensional space faithfully, the points with moderate or large
distances between them in high-dimensional space are placed too far away
from each other in low-dimensional space. Therefore, in the visualization, the points with 
small or moderate distances between them crash together. To solve this problem, 
UNI-SNE \cite{cook2007visualizing} adds a slight repulsion strength to any pair of points in low-dimensional 
space to prevent the points from moving too far away from each
other. Alternatively, t-SNE \cite{maaten2008visualizing} \cite{van2014accelerating} 
uses the Student's $t$-distribution in low-dimensional space, which has a heavy tail compared with the Gaussian distribution used in high-dimensional space.
Kobak et al. \cite{kobak2019heavy} further extended the idea to use the $t$-distribution with one degree of freedom $v$. 
They show that for some data sets, setting $v<1$ would reveal additional
local structure of the data compared with the Student's $t$-distribution with
$v=1$. On the other hand, UMAP uses a curve that is similar to the $t$-distribution but with two
hyperparameters, $a^*$ and $b^*$, in low-dimensional space. The two
parameters are estimated by approximating an exponential function with the
parameter $min\_dist$. By adjusting $min\_dist$, UMAP
governs the appearance of the embedding.

In this paper, we propose using a generalized sigmoid function to model
the distance similarity in both high- and low-dimensional spaces.
In particular, the parameter $b$ is introduced to the generalized
sigmoid function in low-dimensional space, and we can adjust the 
heaviness of the function tail by changing the value of $b$.
We use both simulated and real-world data sets to show that our proposed method
can generate visualization results that are competitive with those of
UMAP. In addition, $b$ can 
be easily adjusted either to reveal the finer cluster structure of the
data or to assist the visualization of data by 
increasing the continuity of neighbors. For some data sets, decreasing
the value of $b$ can provide greater visibility of the intrinsic
structure of the data that might not be
visible using conventional UMAP. Finally, we use domain knowledge 
to demonstrate that the finer subclusters are meaningful.

\section{Methodology}
Let $X=\{ x_1, \ldots, x_N\}$ be $N$ input data in high-dimensional space $R^D$, and let
$d(x_i,x_j)$ be the distance between $x_i$ and $x_j$ under the metric 
$d: X \times X \rightarrow \mathbb{R}_{\geq 0}$. Let $N(x_i)$ be
the \emph{k}-nearest neighbors of $x_i$, and let $\rho_i$ be the shortest distance
from $x_i$ to its \emph{k}-nearest neighbors. Here $\rho_i$ is defined as
\[
\rho_i = \min\{d(x_i, x_j) | d(x_i, x_j) > 0, x_j \in N(x_i),  1 \leq j
\leq k\},  \quad 1 \leq i \leq N
\]

For each $x_i$, we use the sigmoid function with the parameter $\sigma_i$ to
model the distance similarity between points $x_i$ and $x_j$, which is
\[
P_{j|i}=\left \{
\begin{array} {lr}
\frac{1.0}{1.0+ \frac{\max(0, d(x_i, x_j)-\rho_i)}{\sigma_i}}& \quad \mbox{if}\, \, x_j \in N(x_i) \\
0                                               & \quad \mbox{otherwise}
\end{array}
\right.
\]


Using the same approach as UMAP, we can determine the value of $\sigma_i$ through the following equation:
\[
\sum_{j=1}^{k} P_{j|i} = \log_2 (k)
\]

For the simplicity of optimizing the loss function and also extracting more structure information in high-dimensional
space, $P(j|i)$ is symmetrized as
\[
P_{ij}= P_{j|i}+ P_{i|j}-P_{i|j}\odot P_{j|i}
\]
where $\odot$ stands for componentwise multiplication. In addition, let
$P_{N\times N}$ denote the matrix of $P_{ij}$. 

Let $Y=\{y_1,\ldots,y_N\}$ denote the representation of $X$ in low-dimensional space. 
UMAP uses the curve 
\[
Q(i, j)= \frac{1}{(1+a^*\|y_i-y_j\|_2^{2b^*})}
\]
to model the distance distribution in low-dimensional space, where $\|y_i-y_j\|_2$ stands for the distance
between points $y_i$ and $y_j$. The curve is similar to the $t$ distribution with two parameters, $a^*$ and $b^*$, which are
estimated by approximating the following function:
\[
\Phi(y_i, y_j)= \left \{ 
\begin{array} {lr}
1                                          &\mbox{if} \|y_i-y_j\|_2 \leq \mbox{min\_dist} \\
\exp(-(\|y_i-y_j\|_2 - min\_dist))  &\mbox{otherwise}
\end{array}
\right.
\]

The parameter $min\_dist$ is an important parameter in the UMAP
algorithm. It controls how closely the points in
low-dimensional representations are packed together. The larger the
value of $min\_dist$ is, the more the embeddings spread out; the smaller the value of $min\_dist$ is, 
the more likely the data in the same cluster are densely packed.
As shown in our experiments, when the value of $min\_dist$ is smaller than 0.01, further decreasing the value of $min\_dist$ usually
does not change the embeddings much. In order to reveal a finer cluster structure, manually decreasing
the values of parameters $a^*$ and $b^*$ might be needed. However, because
there are two hyperparameters, more experiments might be necessary to find the proper values
of $a^*$ and $b^*$. This is beyond the scope of the paper. 

Here we assume that the membership strength of $y_i$ and $y_j$ can be modeled using a 
generalized sigmoid function \cite{ceriotti2011simplifying}, which can be expressed as
\begin{equation}
Q(i, j)= \frac{1.0}{\left[1.0+(2^{u/v}-1)
\left(\frac{\|y_i-y_j\|_2}{s}\right)^u\right]^{v/u}}
\end{equation}

Let $a=u$ and $b=v/u$. Then Equation (1) can be rewritten as

\begin{equation}
Q(i, j)=
\frac{1.0}{\left[1.0+(2^{1/b}-1)\left(\frac{\|y_i-y_j\|_2}{s}\right)^a\right]^{b}}
\end{equation}

When $b=1$, $Q(i,j)$ becomes
\[
Q(i, j)=
\frac{1.0}{\left[1.0+\left(\frac{\|y_i-y_j\|_2}{s}\right)^a\right]}
\]
This is equivalent to the model of UMAP with $a=2b^*$ and $s^{-a} =a^*$.

Here we consider a simplified version of $Q(i,j)$ by rescaling
$\|y_i-y_j\|_2$ by $s$, so that we have
\begin{equation}
Q(i, j)= \frac{1.0}{\left[1.0+(2^{1/b}-1) \|y_i-y_j\|_2^a\right]^{b}}
\end{equation}
In other words, the solution of Equation (3) will have a scale change by
$s$ compared with the solution of Equation (2).

As with the probabilistic model \cite{tang2016visualizing}, the loss function can be written as

\[
L = -\left(\sum_{(x_i, x_j) \in E}P_{ij}\log Q(i,j) + \sum_{(x_i, x_j)\notin
E}\log(1-Q(i,j))\right)
\]

where $E$ is the collection of points $(x_i, x_j)$ for which either 
$x_i \in N(x_j)$ or $x_j \in N(x_i)$.

Using the negative sampling strategy proposed by \cite{mikolov2013distributed},
the loss function can be further written as 

\[
L = - \left(\sum_{(x_i, x_j) \in E}P_{ij}\log Q(i,j) + \sum_{k=1}^M \log(1-Q(i,k))\right)
\]

where $M$ is the number of negative samples for each vertex $i$. 
The gradient of the loss function can be written as
\[
\frac{\partial L}{\partial y_i} = - \left(\sum_{(i, j) \in E}P_{ij}\frac{\partial{Q(i,j)}/\partial{y_i}}{Q(i,j)}
-\sum_{k=1}^M \frac{\partial{Q(i,k)}/\partial{y_i}}{(1-Q(i,k))}\right)
\]
where 
\[
\partial{Q(i,j)}/\partial{y_i}= -ab\left(1.0+(a^{\frac{1}{b}}-1)\|y_i-y_j\|_2^a\right)^{-b-1}(2^{\frac{1}{b}}-1)\|y_i-y_j\|_2^{a-2}(y_i-y_j)
\]
As with UMAP, the stochastic gradient descent algorithm can be used to optimize the loss function. 
                        
In our experiments, we found that the proper range of $a$ is
$[1,1.5]$. The value of $a$ is less important than the value of $b$, and 
setting $a=1$ generally gives us satisfactory results. So, in our
experiments, we set $a=1$. Because $b$ controls the rate
of the curve approaching 0 and 1, adjusting the value of $b$ can affect
the embeddings in low-dimensional space and the data visualization.
To understand how the family of Equation (3) behaves with varying $b$, we draw plots by setting $a=1$ and $b=0.5, 1, 2, 5, 10$. 
The results are summarized in Figure 1.

\begin{figure}[ht]
  \includegraphics[width=10cm]{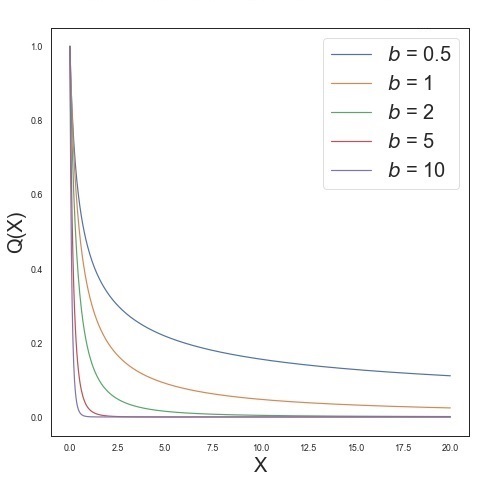}
  \caption{The distance similarity in low-dimensional space is modeled
  by the function in Equation (3). The graph shows the
  curve changes with $a=1$ and $b=0.5, 1, 2, 5$, and $10$.}
  \label{fig:low-dimensional distance}
\end{figure}

We can see from the graph that the smaller the $b$ value is, the more heavy-tailed the curve is. 
The heavy-tail property of the curve can greatly alleviate the crowding
problem when you are embedding high-dimensional data in low-dimensional
space and thus provide the possibility of revealing the finer structure
of the data. 

\section{Experimental Results}
\subsection{Simulated Data}
To study the embeddings with different $b$ values, we randomly generate a 
data set that contains 1,000 data points and 20 dimensions. These data points are evenly distributed among 10 
clusters (each cluster has 100 data points). Within each cluster, the first 50 data points are
randomly sampled from a Gaussian distribution with mean
$\mu_i=5e_i+2.3e_{10+i}$; and the other 50 data points have mean
$\mu_i=5e_i-2.3e_{10+i}$, where $e_i$ is the $i$th basis vector and
$i=1, 2,\ldots,10$. All the data points have covariance $I_{20}$. 
The similar experiment is also considered in \cite{{kobak2019heavy}}. 
In this experiment, we set $k=10$ (10 nearest neighbors), $a=1$, and
$b=0.5, 1, 2$, and $10$. For comparison, we also run UMAP by setting
$k=10$, and $min\_dist=0.001, 0.01, 0.1$, and $1$. For both settings, the initial values of the embedding are set to be 
the eigenvectors of the normalized Laplacian. And 500 epochs are used in the stochastic gradient descent algorithm.
The results are summarized in Figure 2.

\begin{figure}[ht]
  \includegraphics[width=18cm]{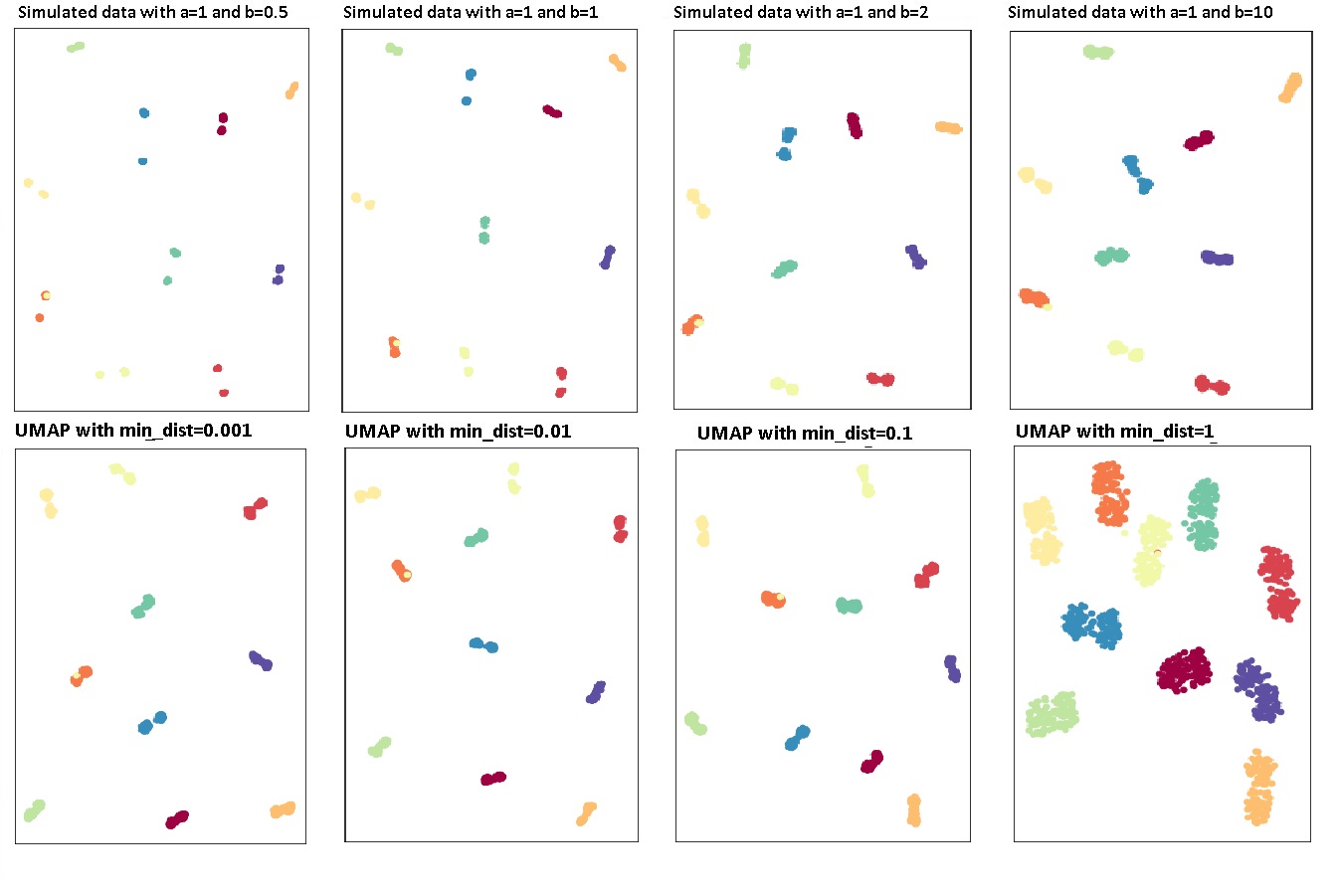}
  \caption{Embeddings comparison using simulated data. For our
  proposed method, the graphs are generated by setting $b=0.5, 1, 2$, and
  $10$, respectively. For UMAP, the graphs are generated by setting
  $min\_dist=0.001, 0.01, 0.1$, and $1$, respectively.}
  \label{fig:Embeddings for simulated data}
\end{figure}

From the setup, we expect that the data will be separated into 10
distinct clusters. Within each big cluster, the data can be classified
into two subclusters or at least have a ``dumbbell'' shape,
because they have different mean values. 

We can see from the embeddings, for our proposed method, that 
all the big clusters are well separated from each other with various
values of $b$. In addition, when $b=0.5$ or $b=1$, the majority of the big
clusters are separated into two isolated subclusters. For the rest of the big
clusters, we can see the dumbbell shape very well. With the value $b$
increasing, the two subclusters within each cluster get closer and
closer and eventually merge into one cluster. However, even with 
$b=10$, we can still see the dumbbell shape within each big cluster.

For UMAP, even with $min\_dist=0.001$, the subclusters within each
big cluster are not well separated. However, we can still see 
the dumbbell shape for some of the clusters, such as the blue, light red, and
yellow clusters. With the increases of $min\_dist$,  the distances between different
clusters decrease. In addition, we further lose the dumbbell shapes 
within several clusters.
We also notice that there is not much difference between the graphs with 
$min\_dist=0.001$ and the graphs with $min\_dist=0.01$. A possible reason is that when
 $min\_dist\leq 0.01$, further 
decreasing the value of $min\_dist$ does not change the values of $a^*$ and
$b^*$ much. 

\subsection{Real Data Sets}

To further test the performance of our proposed method, we apply the
method to the following real data sets with either large or small sample sizes:

\begin{enumerate}

\item MNIST \cite{mnist}: Data set includes 70,000 images of the 
      handwritten digits 0--9. Each image is $28\times 28$ pixels in size. 
\item Fashion-MNIST \cite{xiao2017fashion}: Data set includes 70,000 images of 10
      classes of fashion items (clothing, footwear, and bags). Because the
      images are gray-scale images, $28\times28$ pixels in size, the feature dimension is 784. 
\item Turbofan Engine Degradation Simulation data set \cite{saxena2008turbofan}:
      Engine degradation data are simulated under different combinations
      of operational conditions. In the data set, 21 sensor measurements for 260 engines
      under six operational conditions are recorded until the engine fails. 
      We assume that all the engines operate normally at the beginning of the
      study. There are a total 53,759 observations in the data set. 
\pagebreak      
\item COIL-20 \cite{nene1996columbia}: Data set includes 1,440 gray-scale
images, $28\times 28$ pixels in size, of 20
      objects under 72 rotations spanning 360 degrees. 
\end{enumerate}

For all the data sets, we considered Euclidean distance. 
We set $k=10$ and vary $b$ values from 1, 2, 5, to 10. As with the simulated data,
the initial values of the embedding are set to be 
the eigenvectors of the normalized Laplacian. 
The visualizations are summarized in Figure 3.

\begin{figure}
  \includegraphics[scale=0.52]{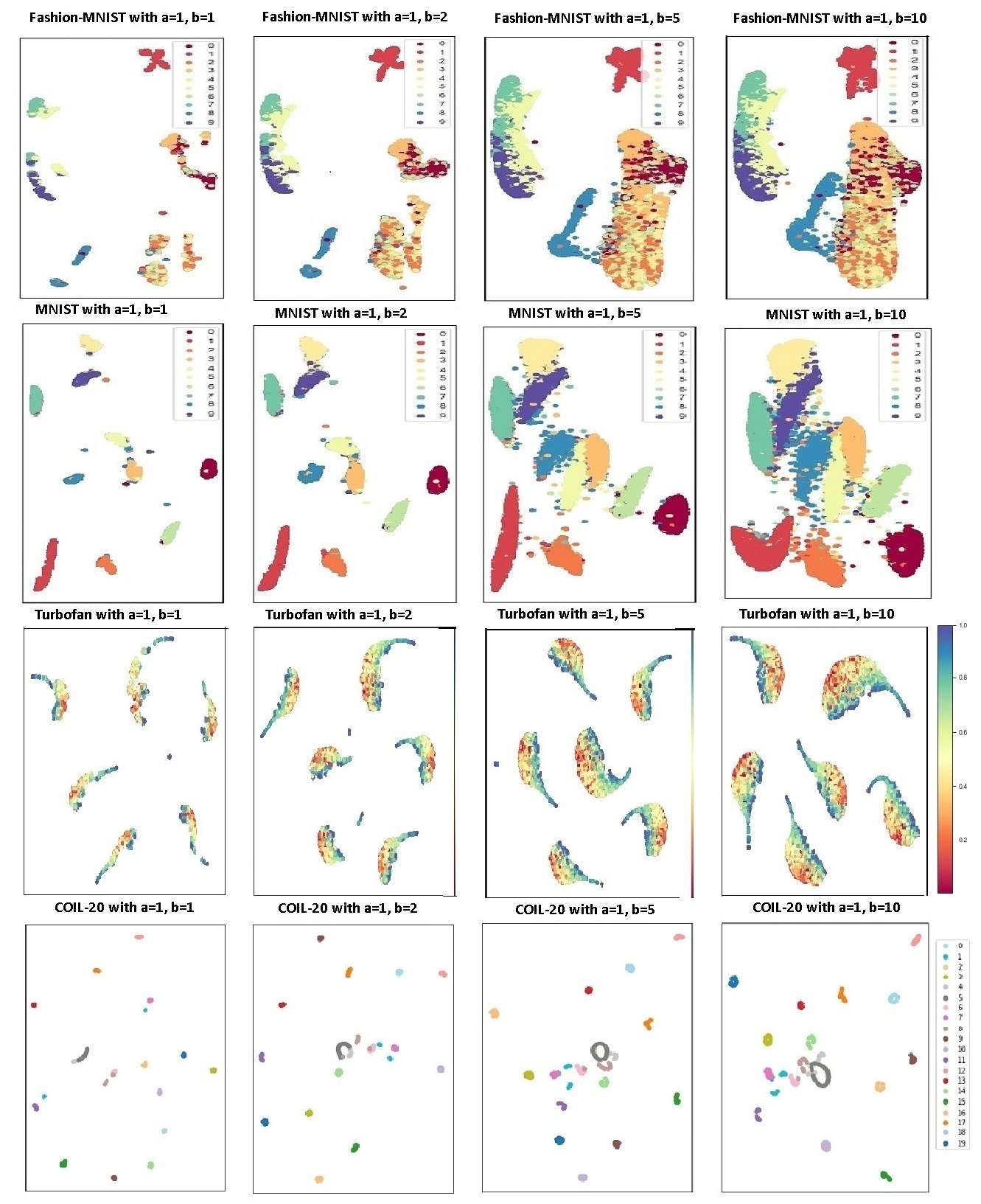}
  \caption{Visualizations of real-world data sets with $b=1, 2, 5$, and $10$, respectively.}
  \label{fig:Embedding for real data sets}
\end{figure}

We can see from the graphs, that when $b=1$, the clusters for each data
set have the biggest separation and the greatest distance. 
For example, all the digits are separated into distinct clusters in the MNIST data set when $b=1$. 
With the increases in the value of $b$, the distances between different digits get smaller and smaller. 
Some digits, such as 4, 7, and 9, and 3, 5, and 8, join together eventually. Based on
the embeddings, we use \emph{k}-means to do classification and find that when
$b=1$ and $b=2$, we get the smallest error, 4.4\%.

For the Fashion-MNIST data set, with various $b$ values, trousers (red) and bags (blue) always 
have the greatest distance from each other. In addition, shoes, bags,
trousers, and other clothes (T-shirts, dresses, pullovers, shirts, and
coats) are well separated when $b=1$ or $b=2$. When $b=5$ and $b=10$,
bags, and other clothes get much closer to each other. They are not distinct clusters anymore. It is also worth pointing out that 
when $b=1$ or $b=2$, we also see a few subclusters that are invisible when $b=5$ or $b=10$. 
For example, the majority of the T-shirts (dark red) and dresses (orange) are separated from coats (yellow), 
pullovers (vermilion), and shirts (light green); sneakers (green), ankle
boots (purple), and sandals (lemon) 
have certain separation as well. In addition, sandals are now separated into two subclusters. 
Among them, one subcluster is close to sneakers and the other is close to ankle boots. 
Furthermore, bags (blue) are separated into two subclusters as well. 

To verify whether these subclusters are meaningful or not, we randomly sampled 100 images from each of the subclusters
that we mentioned earlier and compared the images. We found that the separation of T-shirts and dresses from
other clothing is due to long or short sleeves. For bags, in one
subcluster, the majority of the bags have handles showing at the top of
the image. However, in the other subcluster, either the bags do not have a handle 
or the handle is not showing at the top of the image. In addition, the images of sandals also show that the majority of 
the sandals in one subcluster have middle or high heels, whereas the sandals in the other subcluster 
are relatively flat. These image comparisons clearly show that the 
subclusters revealed by small $b$ values are meaningful. They provide
additional insights into the data structure of Fashion-MNIST. The
comparison results are summarized in Figure 4.  

\begin{figure} 
  \includegraphics[width=13.5cm]{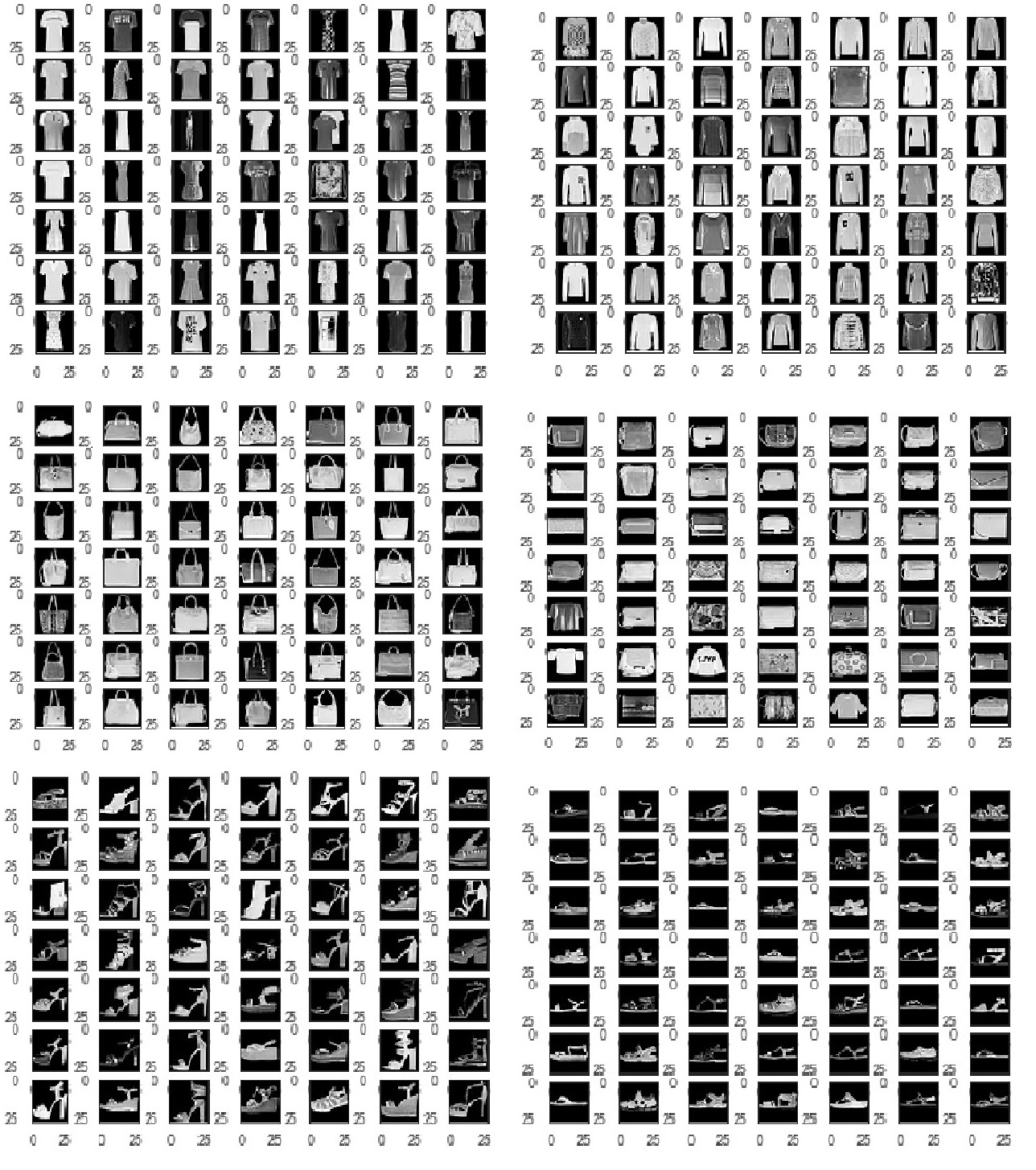}
  \caption{Image comparisons for Fashion-MNIST data. The first row shows the cluster of T-shirts and dresses versus 
  the cluster of coats, pullovers, and shirts. The second row shows the subclusters of bags with and without handles.
  The third row shows the subclusters of sandals with and without heels.}
  \label{fig:image comparisons for fashion-mnist data}
\end{figure}

In the Turbofan Engine Degradation Simulation data set, the flight condition indicator is removed from the data. 
Using the readings only from 21 sensors, our proposed method successfully classified the data into six 
categories with high accuracy. For each cluster, we can see that 
the readings that are taken close to the fault points are approximately on the edge of the embedding (blue color).  
To further investigate the engine degradation process, we remove the
impact of different flight conditions by subtracting 
the average reading measurement for each sensor at each flight condition
and redo the embedding. The resulting embeddings are shown in Figure 5.

\begin{figure}
  \includegraphics[scale=0.4]{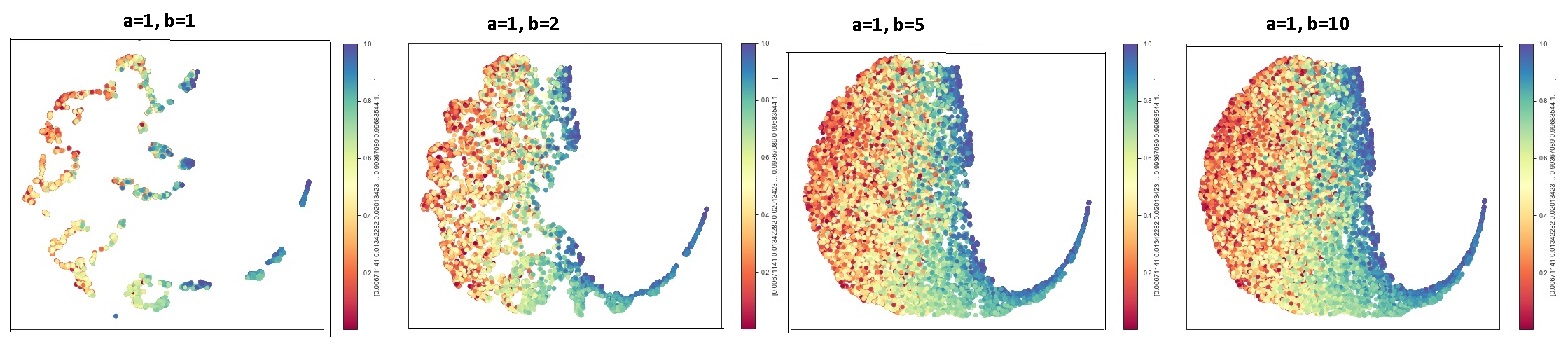}
  \caption{Visualization of Turbofan Engine Degradation Simulation data after removing the impact of different flight conditions.}
  \label{fig 5: Visulization of Turbofan Engine Degradation Simulation Data after removing the impact of different flight conditions.}
\end{figure}

The embeddings clearly show that the sensor readings in the early stage of the study mainly concentrate on one side
of the graph and the readings close to the fault points mainly
concentrate on the other side of the graph with the tail. With a small
value of $b$, we can see that the readings recorded at a similar stage
of an engine's life cycle tend to concentrate
together.

For the COIL-20 data set, with different $b$ values, the majority of the objects are
well separated, except for cars (objects 3, 6, and 19), Anacin, and
Tylenol. With the increases in the value of $b$, the between-class distance
for different objects gets smaller and smaller, which can cause 
problems for data clustering. However, we can see more
clearly the circular structure of each object with a higher value of $b$.
One interesting fact is that we find that object 1 is
classified into three subclusters (light blue color). We randomly
sampled some of the images from each subcluster and found that the 
subclusters are formed mainly according to the direction of the arrow
(downward, upward, and horizontally). Some of the images are displayed in Figure 6.

\begin{figure}
  \includegraphics[scale=0.5]{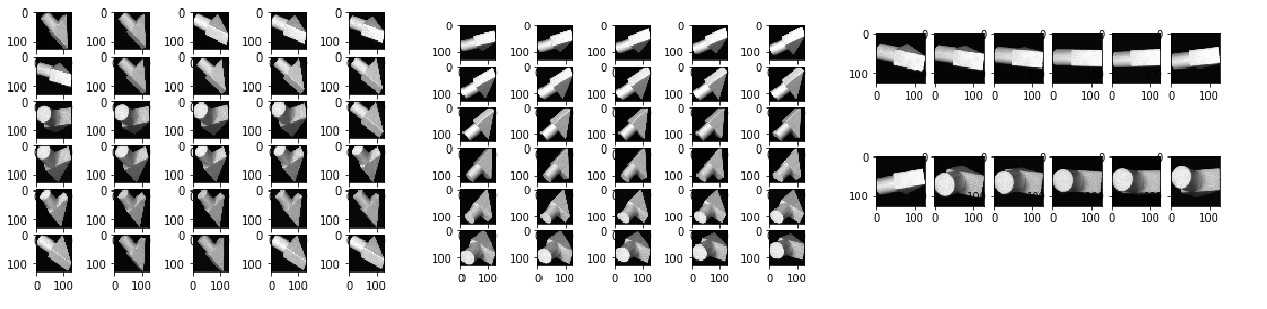}
  \caption{Images for object 1 from three subclusters. The arrows from
  subcluster 1 mainly point downward. The arrows from subcluster 2 mainly
  point upward. The arrows from subcluster 3 mainly point horizontally. }
  \label{fig: arrow image.}
\end{figure}

For comparison, we also run UMAP by setting $k=10$ and $min\_dist$ to 
0.001, 0.01, 0.1, and 1 while keeping other settings the same as in our
proposed method. The results are summarized in Figure 7. From the graph we 
can see that UMAP generates excellent visualization for every data set in general, 
with the majority of the clusters well separated. However, it fails to
separate some clusters that are very similar to each other and  
fails to reveal the subtle subclusters as we see for the Fashion-MNIST
and MNIST data sets with various $min\_dist$ values. It is not sufficient to get a finer
cluster structure by reducing only the value of $min\_dist$. Manually
decreasing the values of the parameters $a^*$ and $b^*$ in UMAP might be needed. 

\begin{figure}
  \includegraphics[scale=0.5]{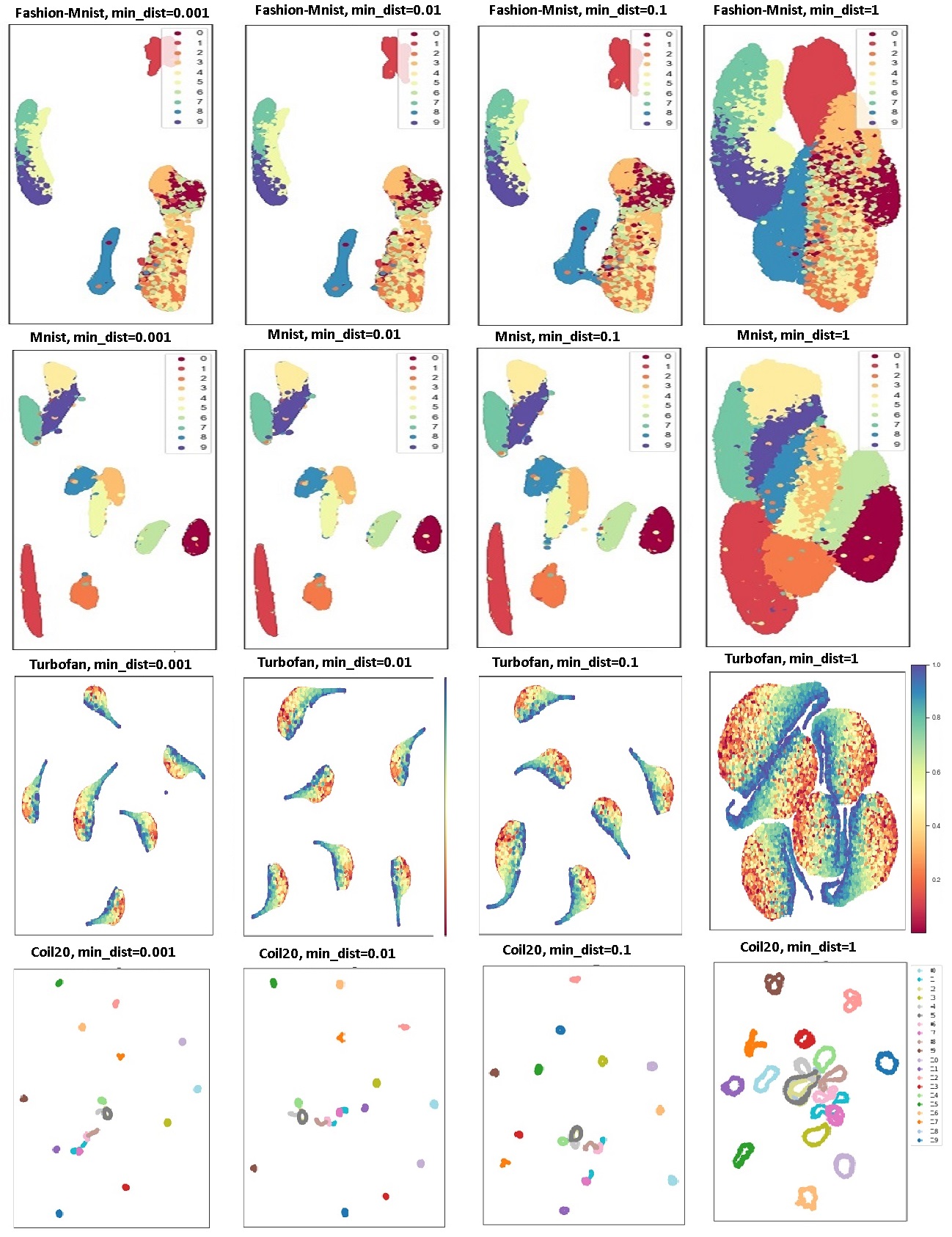}
  \caption{Visualizations of real-world data sets by UMAP by setting $min\_dist=0.001, 0.01, 0.1$, and $1$.}
  \label{fig:Embeddings for real life data sets using UMAP}
\end{figure}

\section{Conclusions}

In this paper, we propose using a generalized sigmoid function to model
the distance similarity in both high-dimensional and low-dimensional spaces.
In particular, the parameter $b$ is introduced to the generalized
sigmoid function in low-dimensional space. By changing the value of $b$,
we can adjust the heaviness of the function tail. 
Using both simulated and real data sets, we show that decreasing the 
value of $b$ can help us reveal the finer cluster structure of the data.
Using visualization and domain knowledge, we show that the
subclusters in the Fashion-MNIST, MNIST, Turbofan Engine Degradation
Simulation, and COIL-20 data sets are meaningful. 

In practice, however, as with the finding in UMAP that a low value of
$min\_dist$ might lead you to spuriously interpret the data structure,
a small value of $b$ might also result in the discovery of some clusters
of random sampling noise. In addition, we learn from the curves with varying $b$ values that the smaller the value $b$ is, the
flatter the tail of the curve is. So embedding convergence might be slower
with a low value of $b$, especially when the sample
size is small, the number of features is high, and
there are many clusters. For example, the COIL-20 data set has only 1,440 images of
20 different objects, but the images have 16,384 features. In order to
get a stable embedding, we increase the number of epochs to 5,000 with small $b$
values. 

In the literature of information visualization, how to assess the quality of graph visualizations is a
long-standing problem. If the purpose of the study is classification or data
exploration, then reducing the value of $b$ properly might give us more
insights into the data structure. However, if the purpose of the study is
to maintain pairwise distance of neighbors in high-dimensional
space, then reducing the value of $b$ can also reduce the continuity of
the neighbors. In practice, there is no unanimous criterion for choosing the $b$
value. We suggest trying out different $b$ values for data exploration
to get a comprehensive understanding of the data.

\section{Acknowledgments}

The authors would like to thank Ed Huddleston, Senior Technical Editor for his assistance in
editing this paper, and Anya McGuirk, Distinguished Research Statistician Developer, and Byron Biggs, Principal
Research Statistician Developer, for their advice and comments.

\newpage
\bibliographystyle{unsrt}  


\end{document}